# A preliminary study of Croatian Language Syllable Networks

K. Ban, I. Ivakić and A. Meštrović*
* Department of Informatics University of Rijeka, Rijeka, Croatia
kristina.ban89@gmail.com, ivan.ivakic@net.hr, amestrovic@inf.uniri.hr

**Abstract** - This paper presents preliminary results of Croatian syllable networks analysis. Syllable network is a network in which nodes are syllables and links between them are constructed according to their connections within words. In this paper we analyze networks of syllables generated from texts collected from the Croatian Wikipedia and Blogs. As a main tool we use complex network analysis methods which provide mechanisms that can reveal new patterns in a language structure. We aim to show that syllable networks have much higher clustering coefficient in comparison to Erdös-Renyi random networks. The results indicate that Croatian syllable networks exhibit certain properties of a small world networks. Furthermore, we compared Croatian syllable networks with Portuguese and Chinese syllable networks and we showed that they have similar properties.

## I. Introduction

Network analysis has become significant method in different research areas such as biology, computer science, economics, sociology, medicine and linguistics. Complex networks are a class of networks that exhibit specific topological features, such as high clustering coefficients, small diameters, power-law degree distribution, community structures, one or several giant components, hierarchical structures, etc. Two important classes of complex networks that can be further differentiated are small-world networks [1, 2, 3] with high clustering as a main property and scale-free networks [4, 5] which can be characterized by power-law degree distribution.

Language can be viewed as a complex network if it is presented as a system of interacting units. Network analysis provides mechanisms that can reveal new patterns in a complex structure and can thus be applied to the study of the patterns in language structures. This, in turn, may contribute to a better understanding of the organization and the structure and evolution of a language.

Network properties of written human languages have already been analyzed in different research studies [6]. Networks based on co-occurrence of words in sentences are analyzed in [7, 8, 9]. The topology of human written language, through a network representation of Orwell's 1984, is presented in [10], while the co-occurrence properties of words in different languages are studied in [11]. All these studies have shown that language networks exhibit properties indicative of small-world networks, e. g. Pemble and Bingol [12] have constructed two complex networks out of Wikipedia English and German corpora and analyze conceptual networks in different languages.

So far, syllable networks have been constructed exclusively for Portuguese [13] and Chinese [14]. In both experiments, syllable networks have a large clustering coefficient and power-law degree distribution, as opposed to the Erdös-Renyi (ER) random networks [15], which have low clustering coefficient and Poisson-like degree distribution. In [13] the syllable network is used to demonstrate that language in itself resembles a living organism, evolving in time and space.

Network analysis of the syllables connections in the words may be of theoretical interest in the domain of phonology, morphology and language topology [16]. Analyses of properties of syllable networks can help in determining the phonetic structure of a language, as well as providing necessary grounds for further linguistic research. Besides theoretical analysis of language, syllable network analysis may be of certain interest in the domain of natural language processing, for speech recognition and speech synthesis. Syllables can be used as acoustic units in automatic speech recognition and as units in text-to-speech systems [17,18,19]. In [18,19] a syllable-based language model is presented and it corresponds to the weighted syllable network.

In this paper we describe experiments with syllable networks for the Croatian language. We constructed four different syllable networks from texts collected Croatian Wikipedia and Blogs. The main goal was to analyze if the Croatian language syllable networks have properties of small-world networks and to analyze if these properties are similar to the properties of Portuguese and Chinese syllable networks. Furthermore, the aim was to compare two different strategies for network construction. As well, we wanted to compare networks from two different text corpora. The presented work is the first attempt to model Croatian syllables as the complex network.

In the second section we present different syllable network construction strategies, text corpora and syllable networks that are constructed from the text. In the third section we describe how to estimate network measures. In the fourth section we present results. In the fifth section we elaborate on the obtained data and provide concluding remarks.

## II. Networks construction

### A. Syllable networks construction strategies

Different strategies can be applied in building syllable networks from text. The idea of a syllable network is to represent syllables as nodes and establish links between them according to their connections within words.

Generally speaking, a syllable network can be either undirected or directed and unweighted or weighted. In a directed syllable network, a directed link indicates the direction of the connection; displaying which syllable (node) is the initial and which syllable (node) is the target. By using a directed network, the successor or the predecessor of an intended syllable can be seen, possibly providing the grounds for further statistical analysis of language structure on the phonetic level. Weighted syllable networks contain information about the number of established links between two syllables, which is again significant in phonetic structure analysis.

A question of how to establish the links between the nodes (syllables) must be discussed. One way is to connect the syllables that belong to the same word (syllable co-occurrence network) and another way is to connect only the neighbour syllables (first-neighbour network). This results in eight different syllable network models. In [13, 14], the network is constructed in a way that two nodes (syllables) are connected if they belong to the same word, making the network undirected and unweighted. This simplified model of a syllable network is constructed in order to study the evolution of the language using phonetic elements [13]. We constructed three networks according to this model.

In our opinion, for some purposes that include natural language processing and linguistic studies, it also makes sense to construct a syllable network of syllables that are direct neighbours in the word. Therefore we additionally constructed and analyzed one directed and weighted neighbour network.

*B. Data*

We analyzed different networks of syllables from different text corpora. The texts used for building the networks are two large corpora. The first corpus is the Croatian Wikipedia. The second corpus contains 3.218 articles collected from different Croatian blogs (including 4 religious and 5 political portals, 6 blog spaces, 3 web-pages with comments and 4 columns from the daily newspapers).

The reason why we have chosen these corpora is because our future work is focused on the text collected from the Web. Another possible approach is to choose a dictionary of Croatian language as a network source. But in [13] it is shown that there is no big difference between syllable networks constructed from the book and syllable network constructed from the dictionary for unweighted networks. A problem we encountered was the Wikipedia corpus containing a certain number of foreign words. This is the reason why the initial network had certain syllables unusual for the Croatian language. Therefore, we examined a filtered network from which all nodes with small degree (meaning that they contain some rare and unusual syllables) were excluded. There is a linguistic difference between the two corpora. The Wikipedia corpus is more formal, so there are more standard words with a pattern in writing. On the other hand, the blog corpus is mostly written in an informal manner, with the use of dialect, slang or abbreviations. However, all of the mentioned texts specifics collected from the web are essential for our future work.

*C. Syllable networks*

We constructed four different networks. Three of them were designed as word co-occurrence syllable networks: the first from the Wikipedia text - $C_W$, the second from the blog text – $C_B$, and the third was devised from both corpora – $C_{WB}$. The fourth network was constructed as a directed and weighted first-neighbour syllable network from the Wikipedia text. The number of nodes and edges for all four networks are displayed in table I.

TABLE I. SYLLABLE NETWORKS CONSTRUCTED

|  | $C_W$ | $C_B$ | $C_{WB}$ | $C_W$-$Dir$ |
|---|---|---|---|---|
| Nodes ($N$) | 4284 | 2000 | 4067 | 4438 |
| Links ($K$) | 170248 | 36202 | 173660 | 33341 |

Network construction is implemented in the Python programming language which contains the NetworkX software package developed for the creation, manipulation, and study of the structure, dynamics, and functions of complex networks [20]. For network visualization we used Gephi software [21]. For the separation of a word into syllables we use syllabification algorithm that is implemented according to the rules described in [16]. The syllabification process is Both corpora where in *txt* file format which made the reading and processing easy and the only problem was the encoding because of our diacritical signs such as č,ć,š etc. The NetworkX module provided us with all the necessary commands to construct a graph and then export it in the desired format.

The co-occurrence syllable network constructed from texts from Wikipedia ($C_w$) visualized using Gephi is shown in Figure 1. The most frequent syllables are pointed out.

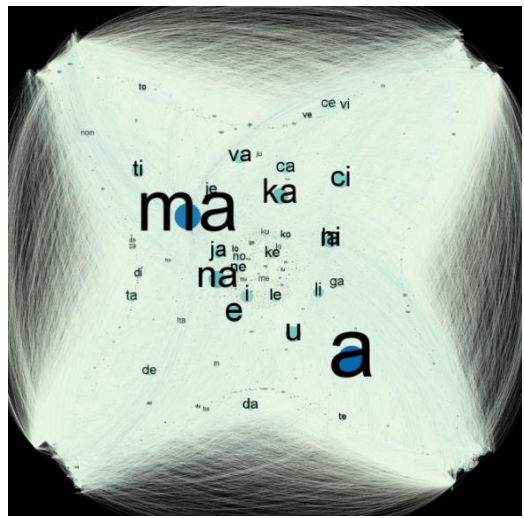

Figure 1. Syllable network from Wikipedia

Another co-occurrence syllable network constructed from blog corpus ($C_B$) is shown in Figure 2. This is a smaller network with smaller number of nodes, but the most frequent nodes (syllables) are similar to the first network, which is discussed in the fourth section.

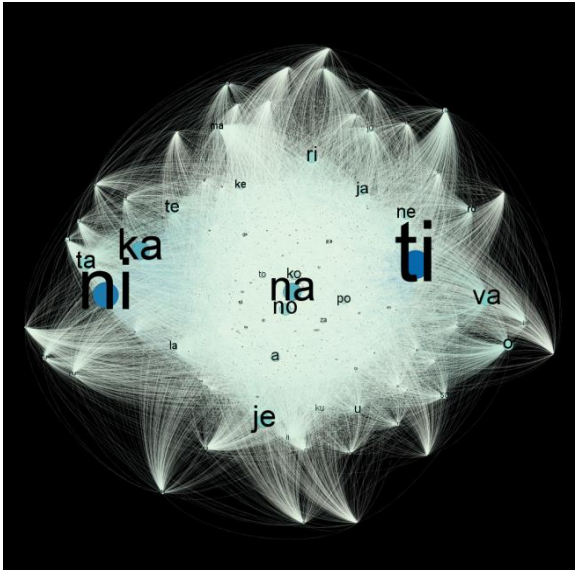

Figure 2. Syllable network from blog corpus

The third network is constructed from both Wikipedia and blog corpora. Syllables with the most connections with other syllables are pointed out and are almost the same as in the first network.

The fourth network is constructed as a directed and weighted network of first neighbor syllables from words that appear in the texts of the Croatian Wikipedia. The idea was to compare this network to the other three networks and to see if it had potential in phonetic structure analysis.

## III. THE NETWORK STRUCTURE ANALYSIS

In this section we explain the most important measures for network analysis. Every network has an $N$ number of nodes and $K$ links.

The degree of a node $i$ is the number of connections of the node and is denoted by $k_i$. Thus, the average degree of the network is:

$$<k> = \frac{2K}{N}$$

For every two connected nodes $i$ and $j$ the number of links lying on the path between them is denoted as $d_{ij}$, therefore the average distance of a node $i$ from all other nodes is:

$$d_i = \frac{\sum_j d_{ij}}{N}$$

From where we easily obtain the average path distance $L$ as the average value of $di$ of all nodes:

$$L = \frac{\sum_i d_i}{N}$$

and the maximum distance results in the network diameter, $D$:

$$D = \max_i d_i.$$

The clustering coefficient is described as a presence of connections between the nearest neighbours of a node. The clustering coefficient $C_i$ of a node $i$ is defined as a ratio between the number of edges $E_i$ that actually exist among these $k_i$ and the total possible number of edges:

$$C_i = \frac{2E_i}{k(k-1)}$$

The average clustering of a network $C$ is the average value of the clustering coefficient of all the nodes:

$$C = \frac{\sum_i C_i}{N}$$

The main property of small-world networks is that the distance between two random nodes grows proportionally to the logarithm of the number of nodes. Therefore, small-world networks tend to have small diameter and short average distance which is the property of random ER networks. Another important property is the high clustering coefficient in comparison to random ER networks. Furthermore, for complex networks it's typically a power-law degree distribution.

## IV. RESULTS

One of our objectives in this experiment is to see if constructed syllable networks of the Croatian language have properties of small–world networks. Small-world properties have already been proven for syllable networks of the Portuguese and the Chinese language; therefore we expected to find similar results for the Croatian language. For the purpose of comparing constructed networks with random networks, ER networks with the same number of nodes and edges have been constructed and all the important properties have been analyzed.

Using Gephi we filtered the networks and determined the average degree $<k>$, diameter $D$, average distance $L$, average clustering coefficient $C$ and some other network values. The correspondent values of these coefficients are shown in table II.

TABLE II. ESTIMATED NETWORK MEASURES FOR CO-OCCUERNCE SYLLABLE NETWORKS

|  | $C_W$ | $ER_W$ | $C_B$ | $ER_B$ | $C_{WB}$ | $ER_{WB}$ |
|---|---|---|---|---|---|---|
| N | 4284 | 4284 | 2000 | 2000 | 4067 | 4067 |
| <k> | 39.74 | 39.74 | 18.1 | 18.1 | 42.7 | 42.7 |
| D | 4 | 3 | 4 | 3 | 3 | 3 |
| L | 2.151 | 2.209 | 2.310 | 2.489 | 2.113 | 2,143 |
| C | 0.691 | 0.017 | 0.687 | 0.016 | 0.690 | 0.021 |

The results show that all three co-occurrence syllable networks have a small diameter and average path distance. Furthermore, for all three networks it holds $<k> \ll N$ which shows that syllable networks are sparse as it is expected for complex networks. In comparison to the ER networks with the same number of nodes and edges these networks show a high clustering coefficient: $C(C_w) \approx 40C(ER_w); C(C_B) \approx 42C(ER_B); C(C_{WB}) \approx 33C(ER_{WB})$.

All these results lead to a conclusion that co-occurrence syllable networks of Croatian language exhibit small world network properties.

We compared our results with the results obtained for the Portuguese and Chinese languages and concluded that there is a similarity between these families of syllable networks. Syllable networks of the Portuguese and of the Croatian language are similar in size, are both sparse, have a small diameter, small size of average path length and both have a high clustering coefficient. Syllable networks of the Chinese have different sizes, but the properties show that these are also small-world networks.

The results of the fourth, weighted and directed first-neighbour syllable network[1] analysis are shown in table III. Although $C$ for the first-neighbour syllable network was smaller than in the co-occurrence syllable networks, in comparison with the random network, it was still about 30 times larger than random network $C$. These values indicate that the first-neighbour syllable network may be a small-world network as well, however, more experiments with larger corpora need to be conducted.

TABLE III. ESTIMATED NETWORK MEASURES FOR THE FIRST-NEIGHBOUR SYLLABLE NETWORK

|   | $C_W$ -Dir | $C_W$ -Undir | ER |
|---|---|---|---|
| $N$ | 4438 | 4438 | 4438 |
| $K$ | 33341 | 33341 | 33341 |
| $D$ | 9 | 8 | 5 |
| $C$ | 0.153 | 0.208 | 0.007 |

In these preliminary experiments we did not estimate the degree distributions for syllable networks. However, we did use NetworkX functions to plot degree distributions on the log-log scale and the result that we got for the co-occurrence syllable network is shown in Figure 3. The straight line on log-log scale indicates that a power-law distribution should be tested in further experiments.

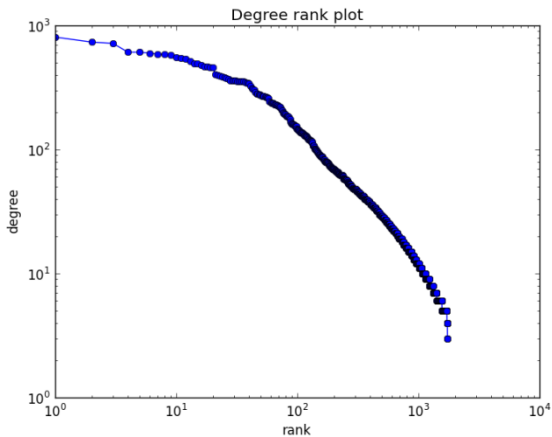

Figure 3. Degree distribution for co-occurrence syllable network

---

[1] For the purpose of some network measures estimation, it was transformed into an undirected, unweighted network.

Small subnetworks were filtered with ten nodes each from all three networks with the highest degree. The results are shown in table IV. It is shown that all three networks have almost the same nodes with the highest degree. This indicates that different corpora do not create significant differences between the networks.

TABLE IV. THE MOST FREQUENT SYLLABLES

| $C_A$ | | $C_W$ | | $C_B$ | |
|---|---|---|---|---|---|
| Syll. | Degree | Syll. | Degree | Syll. | Degree |
| ma | 2299 | ma | 2296 | ma | 836 |
| na | 2166 | na | 2156 | ti | 824 |
| ni | 1937 | ni | 1927 | na | 753 |
| ti | 1918 | ra | 1897 | ni | 741 |
| ra | 1894 | a | 1890 | ka | 627 |
| a | 1860 | ti | 1889 | ci | 626 |
| ne | 1808 | ne | 1792 | ra | 623 |
| ka | 1801 | ka | 1773 | je | 604 |
| o | 1692 | o | 1672 | no | 595 |
| ta | 1682 | ta | 1670 | ne | 593 |

V. CONCLUSION

In this paper we presented different approaches in syllable networks construction. Undirected and unweighted word co-occurrence syllable networks have been already constructed and analyzed for two languages: Portuguese and Chinese. The same syllable networks constructed for the Croatian language (from different corpora) exhibited similar results. The networks contain a high cluster coefficient compared to random networks of the same size and small diameter and average path length. In conclusion, the Croatian language syllable networks have properties of small-world networks.

Another approach was to construct a directed and weighted first-neighbour syllable network for the Croatian language. As far as we know, this is the first time this syllable network construction type has been utilized. The main idea of this approach is to capture more information about the properties of each syllable (the successor, the predecessor and strength of connections with other syllables). It is shown that this kind of network has small-world network properties as well.

These are just preliminary results and there is still a lot of future research to be conducted in this direction. The syllabification algorithm from [16] should be reconsidered and the correctness of the Croatian syllabification should be assessed.

Furthermore, detailed statistical analysis should be performed. The experiment should be repeated with larger corpora such as Croatian literature and dictionaries. However, it is necessary to determine an exact degree distribution for all networks. Our plan is to analyze the network growth and possible communities in the network.